\documentclass[sigconf]{acmart}

\usepackage{tikz}\usetikzlibrary{arrows.meta,positioning,calc}
\usepackage{booktabs}
\usepackage{multirow}

\AtBeginDocument{%
  }

\setcopyright{none}
\renewcommand\footnotetextcopyrightpermission[1]{}
\acmConference[USRW]{Unified Search and Recommendation Workshop, co-located with ACM RecSys 2026}{2026}{Minneapolis, MN, USA}

\begin{document}

\title{Recovering Expert Critic-Sourced Adjacency between Musical Artists from Acoustic Distributions: A Construct-Validity Approach}

\author{Elena Badillo-Goicoechea}
\orcid{0000-0003-1653-1247}
\affiliation{%
  \institution{University of Chicago}
  \city{Chicago}
  \state{Illinois}
  \country{USA}
}
\email{elenabadillog@uchicago.edu}

\author{Fengfeng He}
\orcid{0009-0002-4705-4797}
\affiliation{%
  \institution{University of Chicago}
  \city{Chicago}
  \state{Illinois}
  \country{USA}
}

\renewcommand{\shortauthors}{Badillo-Goicoechea and He}

\begin{abstract}
Music recommendation and discovery systems rely primarily on two kinds of signal: user--item interactions, which fail in the cold-start regime, and intrinsic musical content, which remains available for any recording and is therefore the natural basis for content-based discovery. We argue that a third signal, largely untapped by these systems, is both richer and more principled: \emph{critical adjacency}, the pairwise relation established when an expert music critic explicitly links two artists in long-form prose. Critical adjacency encodes deliberate, articulated judgments about which artists belong together, and prior work introducing this approach established its internal validity---recovering coherent, musicologically interpretable communities under network analysis and matching collaborative filtering in user-satisfaction simulations---while requiring no user data~\cite{badillo-goicoechea-hdsr}. What has been missing is external validation: is this critic-sourced relation grounded in the music itself, or merely in social and narrative context? We supply that validation by testing it against acoustic content, reframing the question as one of construct validity.

Representing each artist as an empirical distribution over 80 low-level acoustic descriptors (Essentia) and modeling pairwise proximity via marginal optimal-transport (Wasserstein) distances, we evaluate the degree to which critical adjacency is sonically recoverable under a cold-start, artist-disjoint split. Our ensemble recovers critic-sourced edges with an overall AUC of 0.767. Crucially, recoverability rises monotonically with critical consensus, reaching AUC 0.865 on multi-source attested edges, while single-source mentions account for much of the residual gap.

Furthermore, typological evaluations align with sociological models of genre: tightly bounded, scene-based genres exhibit higher sonic recoverability than broad, retrospective industry umbrella terms. These findings demonstrate that audio distributions provide an effective physical instrument for decomposing critical discourse: multi-critic consensus identifies a reproducible ``sonic core,'' whereas single-source edges mark a ``sociological remainder,'' potentially driven by narrative positioning, subcultural context, and canonical placement. This work offers both a scalable discovery mechanism for cold-start recommendation and a sociologically grounded approach to MIR and MRS research by incorporating insight from the sociology of popular music.
\end{abstract}


\maketitle

\section{Introduction}

In music recommendation and discovery, interaction-based signals like streaming counts and playlist co-occurrences degrade when items lack historical interaction data, a fundamental boundary known as the cold-start problem~\cite{schein2002coldstart, vandenoord2013deep}. Content-based methods bridge this gap by deriving similarity directly from intrinsic item properties rather than accumulated consumer behavior. Within standard recommender system taxonomies~\cite{javed2021review, knees2013survey, schedl2014mir}, techniques split across collaborative filtering (user logs), content-based methods (raw audio features), and context-based or cultural methods (text, tags, co-occurrences).

When evaluating content-based audio representations, Music Information Retrieval (MIR) research often measures acoustic features against high-level cultural categories or editorial targets. However, interpreting these context-based signals requires careful theoretical grounding. In this work, we analyze critical adjacency, defined as the pairwise relation established whenever a professional music critic co-mentions two artists within long-form journalism, such as album reviews. Unlike the interaction and content signals that current systems rely on, critical adjacency is a deliberate, expert-articulated judgment about which artists belong together, and it remains largely untapped by recommendation and discovery systems. Prior work introducing this approach established that these critic-sourced relations form a coherent, internally valid artist graph---validated through community-detection analysis of its network structure and through simulation of recommendation quality~\cite{badillo-goicoechea-hdsr}; what has not been established is whether that graph is grounded in the music itself.

Rather than treating critical co-mentions as a direct proxy for acoustic similarity, we reframe this relationship as a construct-validity problem~\cite{sturm2023validity}. By establishing purely content-based, distributional representations of audio as an independent physical baseline, we pose a fundamental question: whether critical adjacency is a valid measure for musical similarity. To answer that, we analyze to what extent critical adjacency is explained by acoustic characteristics, and how much of it reflects non-acoustic social context.

Cultural sociology provides essential context for understanding this distinction. Music reviews are inherently dual-natured: critics function as institutional gatekeepers~\cite{hirsch1972} and cultural intermediaries engaged in acts of consecration and symbolic position-taking~\cite{bourdieu1984, bourdieu1993}. When a critic co-mentions two artists, they construct aesthetic meaning. That connection may stem from shared acoustic characteristics (timbre, tempo, production style), but it may equally reflect shared subcultural scenes, political stances, visual personas, or canonical lineage~\cite{frith1996, lindberg2005}.

Testing critic graphs against low-level audio features serves two main functions:
\begin{enumerate}
\item \textbf{Validating Data Quality:} Album reviews capture deliberate, expert-articulated relationships that provide richer structural signal than unmoderated crowd tags or raw web co-occurrences~\cite{cohen2000web}.
\item \textbf{Measuring Construct Validity:} Because critical discourse is shaped by complex social and narrative forces, audio features provide an objective baseline to determine how much of critical adjacency is sonically grounded versus how much forms a non-acoustic ``sociological remainder.''
\end{enumerate}

A prerequisite for both is a content-based measure of musical similarity that is faithful to how an artist actually sounds. Our second contribution is such a measure: rather than reducing an artist to a mean feature vector, we represent each artist as a \emph{distribution} over their tracks and define similarity as the vector of per-feature Wasserstein (optimal-transport) distances between two artists' distributions. This distributional formulation captures differences in the full shape of an artist's sound (its spread and multimodality), not only its average, and it is the predictor on which every result below rests.

To evaluate this, we model each artist as a continuous distribution over 80 low-level acoustic features, computing pairwise proximity using marginal Wasserstein (optimal transport) distances~\cite{peyre2019computational, villani2009optimal}. We then train a stacked ensemble learner (Super Learner)~\cite{vanderlaan2007superlearner} to predict critical adjacency across a review-derived graph of 19{,}059 artists under an artist-disjoint, cold-start evaluation.

Our findings demonstrate that acoustic distributions capture substantial signal present in critical judgments. More importantly, error patterns illuminate the underlying construct: as critical consensus grows from single mentions to multi-source agreement, audio recoverability rises monotonically from AUC 0.767 to 0.865. Multi-critic convergence isolates a reproducible sonic core, while single-source edges mark the sociological remainder where criticism operates primarily through narrative framing and canonical placement.

\section{Related Work}

\subsection{MIR Construct Validity}
The clearest theoretical framework for reframing the paper is the validity typology applied to Music Information Retrieval~\cite{sturm2023validity}. They distinguish statistical conclusion, internal, construct, and external validity, and argue that MIR studies often conflate these forms. In particular, they question the common assumption that genre-classification accuracy provides evidence of music similarity. This framework provides the terminology needed to state explicitly that the present study evaluates the construct validity of using critic co-mentions as an operationalization of musical similarity rather than treating that assumption as implicit.

Flexer \& Grill~\cite{flexer2016} provide important context for interpreting the observed AUC values. They show that agreement between human listeners on music similarity is only moderate, while agreement by the same listener across repeated evaluations is only slightly higher. These findings suggest that part of the gap between the observed AUC values of 0.767 to 0.865 and perfect classification may reflect disagreement within the critic-based similarity construct itself rather than limitations of the audio features or classification models.

Wiggins~\cite{wiggins2009} presents the opposing position that musical meaning is inherently subjective and context dependent, making the notion of a fixed semantic ground problematic. The consensus-stratification results reported in Section~\ref{sec:consensus} provide evidence against the strongest version of this argument. Critic relationships supported by multiple independent sources are more readily recovered from audio features than relationships supported by only a single source, suggesting that stronger consensus corresponds to greater acoustic coherence.

Ellis et al.~\cite{ellis2002} provide the closest precedent for treating musical similarity labels as contestable rather than fixed. They compared several candidate sources of artist similarity, including expert-generated lists, critic co-mentions, and listener behavior, and found limited agreement between them. This supports interpreting the critic network used in the present study as one possible operationalization of similarity rather than an unquestioned ground truth. Berenzweig et al.~\cite{berenzweig2004} provides a natural companion reference for comparisons between acoustic similarity and subjective similarity judgments.

One additional explanation that may warrant investigation is hubness~\cite{radovanovic2009}. In high-dimensional nearest-neighbor spaces, a small number of items can become frequent neighbors regardless of their true similarity. Related works suggest that hub recommendations are often judged by listeners to be less perceptually meaningful. An empirical test would examine whether false-positive neighbor retrieval in the present study is concentrated among a small number of hub artists.

\subsection{Genre as a Limited Proxy for Musical Similarity}
The limitations of genre as a standalone proxy for musical similarity are well documented within both the MIR and the sociological literature. McKay \& Fujinaga~\cite{mckay2006} respond directly to the view that genre classification should be abandoned in favor of more general music similarity research because genre labels are inherently ambiguous and subjective. The existence of this debate shows that concerns about the validity of genre as a classification target have long been recognized within the field.

Empirical evidence also highlights the limitations of genre as a labeling framework. Cross-collection evaluations by Bogdanov et al.~\cite{bogdanov2016} show that genre classification models generalize poorly across datasets, even when the same genre labels are used, because annotators apply labels such as ``Rock'' inconsistently. Seyerlehner et al.~\cite{seyerlehner2010} likewise report human agreement rates ranging from 26\% to 71\% across a 19-genre labeling scheme. These findings provide empirical context for the substantial variation in within-genre AUC observed across genres in Table~\ref{tab:genre}.

The sociological literature offers a complementary explanation for this variation. Lena \& Peterson~\cite{lena2008} identify four broad genre types (avant-garde, scene-based, industry-based, and traditionalist) based on an analysis of 60 forms of American music. Lena~\cite{lena2012} further argues that genres commonly evolve through these forms over time. This framework suggests that genres organized around tightly bounded and internally regulated scenes, such as metal, hip-hop/rap, or electronic music, may exhibit greater sonic coherence than broader industry-based or traditionalist categories such as classical, pop, or alternative rock. Fabbri's~\cite{fabbri1982} account of genre as a socially negotiated system of rules, together with Negus'~\cite{negus1999} critique that these rules are historically dynamic rather than fixed, further supports the view that genre is a social classification rather than a naturally defined acoustic category.

Silver et al.~\cite{silver2016} reach a similar conclusion from a large-scale network analysis of approximately three million musicians. Rather than finding a single, unified genre structure, they identify multiple genre complexes composed of smaller and differently organized communities. Their findings are consistent with the genre heterogeneity observed in the present study and support treating genre as a collection of distinct social formations rather than a single, uniform unit of analysis.

\subsection{Network Approaches and Recommendation Taxonomies}
Network-based approaches in Music Recommender Systems have increasingly moved beyond traditional collaborative filtering by representing relationships as graphs~\cite{badillo-goicoechea-hdsr, mcfee2012}. Most existing network models fall into two broad categories~\cite{knees2013survey, schedl2014mir, vahutre2025}. The first consists of behavioral networks derived from user listening histories, session transitions, and playlist co-occurrences. The second consists of contextual networks built from web pages, microblogs, and other online co-occurrence data. Closest to our use of text, Oramas et al.~\cite{oramas2015semantic} derive artist similarity from a semantic network of entities and relations mined from unstructured music text, showing that text-derived relations capture artist similarity that audio alone does not; our work differs in taking expert-critical co-mention specifically as the relation to be \emph{validated} against audio, rather than combined with it for retrieval.

Each approach has limitations. Behavioral networks depend on user interactions, making them vulnerable to cold-start problems for new or less popular artists while reinforcing existing popularity patterns. Contextual networks constructed from general web data often combine genuine musical relationships with unrelated signals such as viral trends, informal discussion, and search-engine optimization.

The critical adjacency framework represents a different type of network. Instead of relying on user behavior or unfiltered web data, it constructs artist relationships from professional music criticism. Because these links originate from deliberate acts of aesthetic evaluation, the resulting graph provides a structured representation of perceived musical similarity that is less dependent on interaction history or unmoderated online content. This also makes the framework suitable for evaluating cold-start recommendations.

Placing the critic graph within the broader MRS taxonomy~\cite{knees2013survey, schedl2014mir} reveals a difference in how co-occurrence data are defined. Early approaches such as Cohen \& Fan~\cite{cohen2000web} extracted artist relationships from general web searches. Knees \& Schedl~\cite{knees2013survey} later grouped co-occurrence sources into four categories: web pages, microblogs, playlists, and peer-to-peer networks. These categories distinguish sources by platform rather than by authorship, and therefore do not separate expert editorial judgments from consumer-generated activity. Recent reviews~\cite{deldjoo2024, schedl2018} show that music recommender systems increasingly combine item similarity with user context~\cite{adomavicius2010}. Within this setting, evaluating the construct validity of expert-derived similarity provides a way to assess whether one commonly used source of relational information captures meaningful musical relationships.

\subsection{The Sociology of Music Criticism}
The sociology of music criticism provides a theoretical explanation for why critic-based similarity should contain information beyond acoustic resemblance. Rather than treating critical comparisons as neutral descriptions of musical sound, this literature understands them as socially situated acts of cultural evaluation.

Bourdieu~\cite{bourdieu1984, bourdieu1993} provides the central theoretical framework. He distinguishes cultural producers from intermediaries such as critics and describes the process of consecration through which critics confer legitimacy on artists. This relationship is reciprocal: critics often remain associated with artists they championed early, while established artists reinforce the authority of critics who helped construct their reputations. Within this framework, linking two artists in a review is not simply a perceptual judgment about musical similarity; it is also an act of canon formation, field positioning, and cultural evaluation. Critical adjacency should therefore be expected to contain information that extends beyond sonic resemblance alone.

Hirsch~\cite{hirsch1972} complements this perspective by examining the institutional relationship between cultural production and criticism. He describes critics as gatekeepers within the cultural industries and argues that their role depends on maintaining a degree of independence from the producers and labels whose work they evaluate. Hirsch therefore provides a theoretical basis for treating critic relationships as distinct from producer or label promotion.

The historical development of rock criticism further supports this interpretation. Lindberg~\cite{lindberg2005} traces the emergence of music criticism from the specialist press of the 1960s into an established cultural institution with its own conventions, professional identities, and comparative practices. Frith~\cite{frith1996} likewise argues that music criticism operates through socially shared forms of evaluation rather than objective descriptions of musical sound.

This perspective also helps explain why cross-genre artist pairs may remain predictable from audio features. Regev~\cite{regev2013} argues that popular music increasingly operates within a shared framework of aesthetic cosmopolitanism in which critics, musicians, and audiences draw on common evaluative standards across genres. If critics employ a genre-transcending aesthetic framework when comparing artists, cross-genre relationships need not be substantially less coherent than within-genre relationships. This provides a theoretical explanation for the findings reported in Section~\ref{sec:genre}. Bourdieu's framework has also been extended empirically: Corciolani et al.~\cite{corciolani2020} show that critics occupying different positions within the cultural field evaluate artists differently, suggesting that outlet or critic-level metadata could be examined directly as moderators of whether critic similarities are driven more strongly by acoustic characteristics or by broader social processes.

\section{Data}
\label{sec:data}

\paragraph{Critical adjacency graph.} Our target is a graph over musical artists whose edges are derived from music-review sources. Reviews are drawn from a corpus of roughly $61{,}000$ long-form album reviews spanning ten music-criticism outlets (for example Pitchfork, The Guardian, Bandcamp Daily, and NPR), following the corpus construction of prior work~\cite{badillo-goicoechea-hdsr}. A directed edge indicates that critics mention one artist in relation to another in their writing (for example, Brian Eno is mentioned in a review of a David Bowie album such as \emph{Low}, 1977), and the edge weight $a_{ij}$ counts the total number of times artist~$i$ is mentioned across the corpus of artist~$j$'s reviews. After matching to the audio corpus (below) and filtering to artists with at least three tracks, the graph contains 19{,}059 artists. As expected of a critic-sourced graph, edges are sparse (density approximately $8.6\times10^{-4}$) and weakly attested: the median edge weight is one, and roughly four in five artist pairs are linked by a single source.

\paragraph{Audio features.} We use low-level acoustic descriptors computed by the Essentia library and distributed through the AcousticBrainz project, a crowd-sourced corpus in which a single pinned extractor was run on each contributed recording, ensuring cross-contributor comparability. Each recording is described by 80 usable scalar descriptors spanning timbre (MFCC means), timbral variability (MFCC variances), harmony (HPCP), spectral shape, rhythm, tonal features, and loudness. Artists in the critic graph are matched to AcousticBrainz by normalized artist name (case-folded, whitespace- and punctuation-normalized) resolved through MusicBrainz artist identifiers, retaining only artists with at least three matched tracks so that per-artist distributions are estimable. After matching, the corpus covers 2{,}154{,}520 tracks across the 19{,}059 artists, with a median of 42 tracks per artist. Because the Essentia descriptors are computable from any recording, this representation remains available even for artists with no interaction history or critical coverage, which is precisely the cold-start regime we target.

\section{Methodology \& Construct-Validity Framework}
\label{sec:methods}

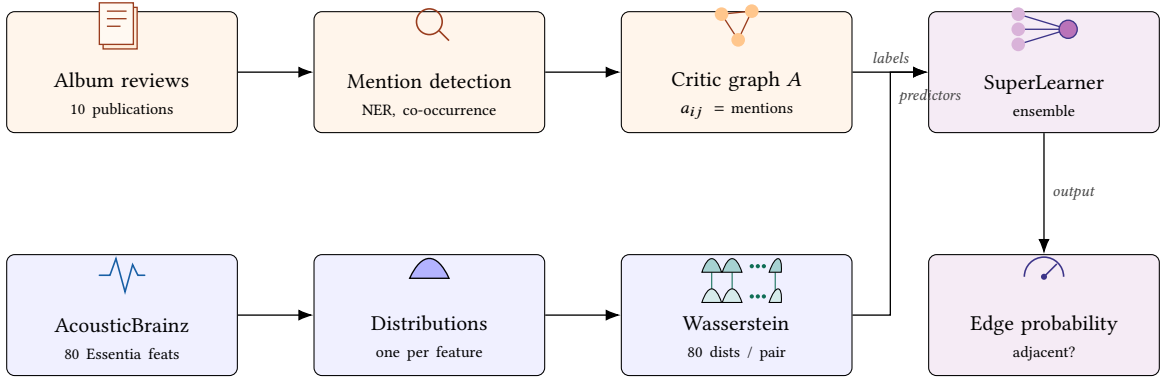
\begin{figure*}[t]
\centering
\definecolor{cri}{RGB}{153,60,29}
\definecolor{aud}{RGB}{24,95,165}
\definecolor{tea}{RGB}{15,110,86}
\definecolor{pur}{RGB}{60,52,137}
\begin{tikzpicture}[
  font=\small,
  box/.style={draw, rounded corners=3pt, minimum height=16mm, text width=27mm, inner ysep=4pt, inner xsep=5pt, align=center, line width=0.4pt},
  audio/.style={box, fill=blue!6},
  critic/.style={box, fill=orange!8},
  learn/.style={box, fill=violet!8},
  lbl/.style={font=\scriptsize\itshape, text=gray!55!black},
  ar/.style={-{Latex[length=2mm]}, line width=0.5pt},
  ard/.style={-{Latex[length=2mm]}, line width=0.5pt, dashed}
]

\node[critic] (reviews) {\\[4.5mm]Album reviews\\{\scriptsize 10 publications}};
\node[critic, right=10mm of reviews] (ner) {\\[4.5mm]Mention detection\\{\scriptsize NER, co-occurrence}};
\node[critic, right=10mm of ner] (graph) {\\[4.5mm]Critic graph $A$\\{\scriptsize $a_{ij}=$ mentions}};
\node[learn, right=10mm of graph] (sl) {\\[4.5mm]SuperLearner\\{\scriptsize ensemble}};
\node[audio, below=16mm of reviews] (tracks) {\\[4.5mm]AcousticBrainz\\{\scriptsize 80 Essentia feats}};
\node[audio, below=16mm of ner] (dist) {\\[4.5mm]Distributions\\{\scriptsize one per feature}};
\node[audio, below=16mm of graph] (wass) {\\[4.5mm]Wasserstein\\{\scriptsize 80 dists / pair}};
\node[learn, below=16mm of sl] (prob) {\\[4.5mm]Edge probability\\{\scriptsize adjacent?}};

\begin{scope}[shift={([yshift=-3mm]reviews.north)}, cri, line width=0.5pt]
  \draw[fill=orange!14] (-2.5mm,-2mm) rectangle (2mm,3.5mm);
  \draw[fill=orange!5]  (-3mm,-1.4mm) rectangle (1.5mm,4.1mm);
  \draw (-2mm,3mm)--(0.6mm,3mm); \draw (-2mm,2mm)--(0.6mm,2mm); \draw (-2mm,1mm)--(-0.2mm,1mm);
\end{scope}
\begin{scope}[shift={([yshift=-3mm]ner.north)}, cri, line width=0.6pt]
  \draw (0,1.6mm) circle (1.7mm);
  \draw (1.3mm,0.4mm)--(2.6mm,-0.9mm);
\end{scope}
\begin{scope}[shift={([yshift=-3mm]graph.north)}, cri, line width=0.5pt]
  \coordinate (n1) at (-2mm,2.6mm); \coordinate (n2) at (2mm,3mm); \coordinate (n3) at (0.2mm,-0.8mm);
  \draw (n1)--(n2)--(n3)--(n1);
  \fill[orange!45] (n1) circle (0.9mm); \fill[orange!45] (n2) circle (0.9mm); \fill[orange!45] (n3) circle (0.9mm);
\end{scope}
\begin{scope}[shift={([yshift=-2.8mm]tracks.north)}, aud, line width=0.6pt, line cap=round]
  \draw (-3mm,0) -- (-1.6mm,0) -- (-0.8mm,2.2mm) -- (0.4mm,-2.2mm) -- (1.4mm,1.4mm) -- (2.1mm,0) -- (3mm,0);
\end{scope}
\begin{scope}[shift={([yshift=-3.2mm]dist.north)}, line width=0.5pt]
  \draw[fill=blue!30] (-2.6mm,0) .. controls (-1mm,3.5mm) and (1mm,3.5mm) .. (2.6mm,0) -- cycle;
\end{scope}
\begin{scope}[shift={([yshift=-3mm]wass.north)}, line width=0.4pt]
  \draw[tea] (-3.3mm,0.6mm)--(-3.3mm,-3.4mm);
  \draw[tea] (-0.6mm,0.6mm)--(-0.6mm,-3.4mm);
  \draw[tea] (5.1mm,0.6mm)--(5.1mm,-3.4mm);
  \draw[fill=teal!35] (-4.6mm,0.6mm) .. controls (-3.8mm,3mm) and (-2.8mm,3mm) .. (-2mm,0.6mm) -- cycle;
  \draw[fill=teal!35] (-1.9mm,0.6mm) .. controls (-1.1mm,3mm) and (-0.1mm,3mm) .. (0.7mm,0.6mm) -- cycle;
  \fill[tea] (2mm,1.4mm) circle (0.28mm) (2.8mm,1.4mm) circle (0.28mm) (3.6mm,1.4mm) circle (0.28mm);
  \draw[fill=teal!35] (4.3mm,0.6mm) .. controls (5.1mm,3mm) and (6.1mm,3mm) .. (5.9mm,0.6mm) -- cycle;
  \draw[fill=teal!16] (-4.6mm,-3.4mm) .. controls (-3.8mm,-1mm) and (-2.8mm,-1mm) .. (-2mm,-3.4mm) -- cycle;
  \draw[fill=teal!16] (-1.9mm,-3.4mm) .. controls (-1.1mm,-1mm) and (-0.1mm,-1mm) .. (0.7mm,-3.4mm) -- cycle;
  \fill[tea] (2mm,-2.6mm) circle (0.28mm) (2.8mm,-2.6mm) circle (0.28mm) (3.6mm,-2.6mm) circle (0.28mm);
  \draw[fill=teal!16] (4.3mm,-3.4mm) .. controls (5.1mm,-1mm) and (6.1mm,-1mm) .. (5.9mm,-3.4mm) -- cycle;
\end{scope}
\begin{scope}[shift={([yshift=-3mm]sl.north)}, pur, line width=0.5pt]
  \coordinate (b1) at (-3.4mm,2.6mm); \coordinate (b2) at (-3.4mm,0.6mm); \coordinate (b3) at (-3.4mm,-1.4mm);
  \coordinate (out) at (3.2mm,0.6mm);
  \draw (b1)--(out); \draw (b2)--(out); \draw (b3)--(out);
  \fill[violet!30] (b1) circle (0.9mm); \fill[violet!30] (b2) circle (0.9mm); \fill[violet!30] (b3) circle (0.9mm);
  \draw[fill=violet!55] (out) circle (1.2mm);
\end{scope}
\begin{scope}[shift={([yshift=-3mm]prob.north)}, pur, line width=0.6pt]
  \draw (-2.6mm,0) arc (180:0:2.6mm);
  \draw (0,0)--(1.7mm,1.7mm);
  \fill (0,0) circle (0.4mm);
\end{scope}

\draw[ar] (reviews) -- (ner);
\draw[ar] (ner) -- (graph);
\draw[ar] (graph) -- (sl) node[lbl, midway, above] {labels};
\draw[ar] (tracks) -- (dist);
\draw[ar] (dist) -- (wass);
\draw[ar] (wass.east) -- ++(5mm,0) |- (sl.west) node[lbl, pos=0.45, right] {predictors};
\draw[ar] (sl) -- (prob) node[lbl, midway, right] {output};

\end{tikzpicture}
\caption{\normalfont Solution architecture. The critic stream (top) builds the directed adjacency graph~$A$ from album-review co-mentions, where $a_{ij}$ counts mentions of artist~$i$ in artist~$j$'s review corpus. The audio stream (bottom) represents each artist as an empirical distribution over 80 Essentia descriptors and compares artist pairs by marginal Wasserstein distances. The two streams converge at the learner: the audio distances are the predictors and the critic graph supplies the supervision, yielding a probability of critical adjacency for any artist pair.}
\Description{A two-stream pipeline diagram. The top (critic) stream flows from album reviews to named-entity mention detection to a weighted critic co-mention graph. The bottom (audio) stream flows from AcousticBrainz Essentia features to per-feature distributions to an 80-dimensional per-pair Wasserstein distance vector. Both streams converge at a SuperLearner ensemble: the audio distances are predictors and the critic graph supplies edge labels, producing a predicted probability that any artist pair is critically adjacent.}
\label{fig:arch}
\end{figure*}

The methodology pipeline (Figure~\ref{fig:arch}) integrates an audio content layer of predictor variables with a context data layer of targets to model critic-sourced artist connections under a cold-start framework.

\paragraph{Audio content layer.} From the 80 Essentia descriptors (Section~\ref{sec:data}), we treat each artist as an empirical distribution over their tracks in the 80-dimensional feature space rather than reducing them to a mean vector: two artists may share an average sound yet differ in how widely their catalogue ranges. This motivates our proposed similarity measure. For each of the 80 features we compute the one-dimensional Wasserstein (optimal-transport) distance between the two artists' empirical distributions of that feature, in closed form as the $L^1$ distance between their quantile functions. Stacking these gives an 80-dimensional \emph{distributional distance vector} that summarizes how two artists differ across the full acoustic spectrum, feature by feature and in distributional shape rather than in mean alone. This vector is the sole predictor used throughout.\footnote{Our aim is not to establish the Wasserstein formulation as an optimal or best-performing audio representation, and we therefore do not treat the choice of acoustic representation as an empirical question to be adjudicated against simpler baselines. We adopt the full marginal precisely because it is, by construction, at least as informative as its mean: a per-feature distribution encodes an artist's catalogue spread, skew, and multimodality, all of which the mean discards. Our object of validation is the \emph{critic} signal, and this calls for as faithful an acoustic characterization as is available, rather than a deliberately impoverished one; a representation that summarizes an artist by feature means alone would validate the critic graph against strictly less acoustic information, weakening exactly the independent baseline the construct-validity argument requires.}

\paragraph{Contextual target layer.} The prediction target is the critic adjacency graph of Section~\ref{sec:data}, constructed by extracting pairwise artist co-mentions from the review corpus via Named Entity Recognition. Entity extraction follows the pipeline of prior work~\cite{badillo-goicoechea-hdsr}: candidate artist mentions are identified with an NLTK/TextBlob named-entity pass and resolved against the roster of reviewed artists, so that recognized entities are constrained to a known artist vocabulary rather than open-domain names. Although the underlying graph is directed---the number of times artist~$i$ is mentioned in artist~$j$'s reviews need not equal the reverse---our predictor, the marginal Wasserstein distance between two artists' acoustic distributions, is symmetric by construction. For this work we therefore collapse the graph to its undirected form, defining an (undirected) edge between two artists whenever a co-mention is attested in either direction and setting its weight to the total co-mention count across both directions; the asymmetry itself is a promising target for future, direction-aware modeling. We predict the presence of a critic-sourced edge between two artists from their acoustic distributional distance vector alone.

\paragraph{Cold-start model training.} A Super Learner stacked ensemble~\cite{vanderlaan2007superlearner} (over linear models, regularized regression, random forests, gradient-boosted trees, and a nearest-neighbor learner) maps the 80-dimensional distance vector to the probability of a critic-sourced edge. Positives are all attested edges; negatives are sampled from non-edges at a 3:1 ratio (case-control sampling). We evaluate under an \emph{artist-disjoint} split: artists are partitioned into training and test sets, the model trains only on pairs internal to the training set and is tested only on pairs internal to the test set, so no test artist is ever observed in training. This is the cold-start condition. Predictive performance is measured by Area Under the ROC Curve (AUC), with bootstrap 95\% confidence intervals.

\section{Results}

\subsection{Construct Validity via Critical Consensus}
\label{sec:consensus}

\begin{figure}[t]
\centering
\refstepcounter{figure}\label{fig:roc}
{\small\textbf{Figure \thefigure: AUC-ROC for recovering critic-sourced edges from acoustic distributions}}\\[2pt]
\includegraphics[width=0.88\linewidth,trim=4pt 120pt 6pt 10pt,clip]{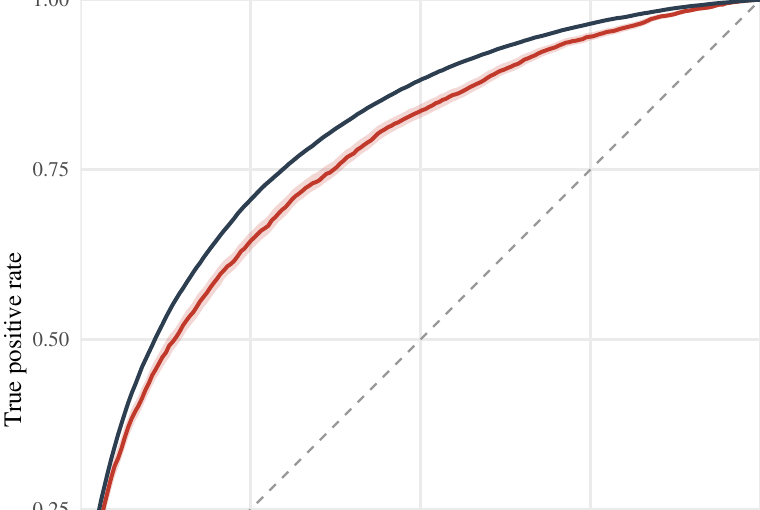}
\Description{Two ROC curves for recovering critic-sourced edges from acoustic distributions. The pair-random cross-validation curve reaches AUC 0.803 and the stricter artist-disjoint cold-start curve reaches AUC 0.767; both rise well above the diagonal chance line, with the pair-random curve slightly higher. A marked operating point on the artist-disjoint curve corresponds to sensitivity 0.69 and specificity 0.71.}
\caption*{\small Note: Pair-random cross-validation (AUC 0.803, 95\% CI 0.800--0.805) and the stricter artist-disjoint, cold-start evaluation (AUC 0.767, 95\% CI 0.761--0.775), where neither artist in a test pair appears in training. The annotated operating point is the Youden-optimal threshold (0.25) on the artist-disjoint curve (sensitivity 0.69, specificity 0.71). Bands are bootstrap 95\% confidence intervals; the pair-random band is narrower than the plotted line.}
\end{figure}

Across the artist-disjoint test split, the ensemble model recovers critical adjacency with an overall AUC of 0.767 (95\% CI 0.761--0.775; Figure~\ref{fig:roc}). Under the less stringent pair-random cross-validation, where an artist may appear in both training and test folds, AUC is 0.803 (95\% CI 0.800--0.805); the model is well calibrated (Brier 0.142 against 0.188 for an intercept-only baseline). To test our construct-validity hypothesis, we evaluate model performance as a function of critical attestation: the number of co-mentions establishing the edge.

The model's performance exhibits a strong upward progression as attestation increases (Table~\ref{tab:consensus}): AUC 0.767 at weight $\geq 1$, 0.788 at $\geq 2$, 0.815 at $\geq 3$, and 0.865 at $\geq 5$.

\begin{table}[t]
  \caption{Audio recoverability rises with critical attestation (artist-disjoint test set).}
  \label{tab:consensus}
  \begin{tabular}{lc}
    \toprule
    Edge attestation & AUC \\
    \midrule
    All edges ($\geq 1$ mention) & 0.767 \\
    $\geq 2$ mentions & 0.788 \\
    $\geq 3$ mentions & 0.815 \\
    $\geq 5$ mentions (higher consensus) & 0.865 \\
    \bottomrule
  \end{tabular}
\end{table}

When an edge is established via higher critical consensus, it exhibits high audio recoverability, reaching AUC 0.865. This convergence isolates a reproducible ``sonic core'' characterized by objective acoustic traits such as shared timbre, tempo, and rhythmic structure, indicating that repeated expert co-mentions strongly reflect physical musical similarity. Conversely, single-source or long-tail edges display significantly lower audio recoverability. Rather than merely indicating model failure, these single-mention links capture a ``sociological remainder'': a discursive space in which critics construct artist relationships using non-acoustic drivers such as shared personas, subcultural narratives, political positioning, and canonical placement. Human inter-rater agreement on music similarity has natural upper bounds~\cite{flexer2016}; the AUC of 0.865 on consensus edges reflects a high degree of acoustic alignment given the inherent variance in human evaluative judgments.

\subsection{Signal Across Genre Typologies}
\label{sec:genre}

To evaluate whether acoustic features capture fine-grained stylistic relationships beyond broad category matching, we examine model behavior within exogenously assigned genres (labels drawn from a controlled vocabulary independent of the audio and of the review sources). If the model were merely a genre detector, discrimination should collapse within a genre, where genre is held constant. It does not: within every adequately powered genre, AUC remains high (Table~\ref{tab:genre}).

\begin{table}[t]
  \caption{Within-genre artist-disjoint AUC: discrimination on test pairs whose artists share an exogenously assigned genre. Bootstrap 95\% CIs. Buckets below 100 pairs are shown for completeness and not interpreted.}
  \label{tab:genre}
  \begin{tabular}{lrcc}
    \toprule
    Genre & Pairs & Prev. & AUC (95\% CI) \\
    \midrule
    Hip-Hop/Rap      & 331 & 0.82 & 0.80 (0.73--0.86) \\
    Electronic       & 273 & 0.59 & 0.80 (0.74--0.85) \\
    Metal            & 251 & 0.53 & 0.79 (0.74--0.85) \\
    Indie            & 132 & 0.55 & 0.81 (0.73--0.88) \\
    Classic Rock     & 112 & 0.66 & 0.71 (0.60--0.80) \\
    Alternative Rock & 109 & 0.58 & 0.72 (0.62--0.81) \\
    \midrule
    \multicolumn{4}{l}{\emph{Underpowered ($<100$ pairs), not interpreted:}} \\
    Country/Folk     &  84 & 0.63 & 0.72 (0.60--0.83) \\
    Pop              &  79 & 0.71 & 0.68 (0.54--0.81) \\
    Jazz             &  75 & 0.71 & 0.80 (0.66--0.91) \\
    R\&B             &  66 & 0.89 & 0.95 (0.88--0.99) \\
    Punk             &  51 & 0.67 & 0.80 (0.66--0.92) \\
    Classical        &  20 & 0.70 & 0.57 (0.29--0.83) \\
    \bottomrule
  \end{tabular}
\end{table}

In a manner consistent with genre typology~\cite{lena2008}, acoustic recoverability varies systematically across organizational forms. Tightly bounded, scene-based genres (Electronic 0.80, Hip-Hop/Rap 0.80, Metal 0.79, Indie 0.81) show strong within-category recoverability, as critical communities enforce clear acoustic and production conventions~\cite{fabbri1982}. Broad industry-umbrella categories (Classic Rock 0.71, Alternative Rock 0.72, Pop 0.68) exhibit lower acoustic recoverability, as critical co-mentions in these spaces are more frequently driven by historical era, commercial stature, or cultural framing. We note that this pattern is suggestive rather than definitive: several sociologically informative buckets (Jazz, Classical, R\&B) fall below the power threshold and are not interpreted.

We further contrast pairs whose artists share a genre against pairs spanning genres. Same-genre AUC is 0.796 (95\% CI 0.775--0.818), slightly \emph{higher} than cross-genre AUC of 0.765 (95\% CI 0.753--0.777). Were genre the operative cue, same-genre pairs (cue absent) would be hard and cross-genre pairs (cue present) easy; the opposite holds. That cross-genre pairs remain nearly as recoverable is consistent with the aesthetic cosmopolitanism~\cite{regev2013} through which critical connections across traditional boundaries often track underlying acoustic affinities.

\subsection{Robustness to Popularity and Exposure}
\label{sec:popularity}

A natural concern is that recoverability may be confounded by popularity or exposure: well-documented artists accrue more reviews, more mentions, higher graph degree, and often more available tracks, any of which could make their relationships easier to model. We assess this directly by stratifying the artist-disjoint test set (24{,}651 pairs drawn only from the 3{,}811 held-out artists) along three exposure proxies: critic-graph \emph{degree} (number of co-mention neighbours), total \emph{mention count} (summed co-mention weight), and \emph{track count} (recordings entering the Wasserstein vectors). Each pair is assigned the \emph{minimum} proxy value across its two artists, so a pair counts as low-exposure whenever \emph{either} artist is sparsely documented; pairs are then partitioned into equal-frequency bins and AUC is recomputed within each bin from the already-trained model, with no refitting (Table~\ref{tab:popularity}).

Recoverability is essentially invariant to degree (0.717 in the least-connected bin vs.\ 0.730 in the most-connected) and to mention count (0.718 vs.\ 0.733), and the least-exposed bins contain the large majority of test pairs, so the signal holds precisely where the critic graph is sparsest. Track count shows the only appreciable gradient (0.679 for artists with 3--11 tracks, rising to 0.748 above 53), which we read as a measurement-precision effect---an artist's acoustic distribution is estimated more accurately from more tracks---rather than a popularity confound, and recoverability remains well above chance even in the most sparsely-sampled bin. Together these results indicate that acoustic recovery of critic adjacency is not an artifact of well-documented or popular artists being easier to model.

\begin{table}[t]
  \caption{Artist-disjoint AUC stratified by exposure proxy (minimum over each pair). Bins are equal-frequency; degree and mention count collapse to three populated bins because the critic graph is heavy-tailed (most artists have degree $0$--$1$).}
  \label{tab:popularity}
  \small
  \begin{tabular}{llrc}
    \toprule
    Proxy & Bin range & Pairs & AUC \\
    \midrule
    \multirow{3}{*}{Degree}
      & 0--1 (lowest)   & 15{,}641 & 0.717 \\
      & 2--3            &  3{,}055 & 0.721 \\
      & 4--364 (highest)&  5{,}955 & 0.730 \\
    \midrule
    \multirow{3}{*}{Mentions}
      & 0--1 (lowest)   & 15{,}196 & 0.718 \\
      & 2--4            &  3{,}583 & 0.720 \\
      & 5--508 (highest)&  5{,}872 & 0.733 \\
    \midrule
    \multirow{4}{*}{Tracks}
      & 3--11 (lowest)  &  6{,}356 & 0.679 \\
      & 12--24          &  6{,}178 & 0.715 \\
      & 25--52          &  6{,}047 & 0.712 \\
      & 53--2614 (highest) & 6{,}070 & 0.748 \\
    \bottomrule
  \end{tabular}
\end{table}

\section{Discussion \& Future Directions}

By framing critical adjacency as a construct-validity problem, this study demonstrates that music reviews integrate two distinct layers of signal: a sonically recoverable core supported by higher critical consensus, and a non-acoustic sociological remainder. Underpinning this analysis is a methodological contribution of independent interest: a distributional similarity measure that represents each artist as a distribution over acoustic features and compares artists by their per-feature Wasserstein distances. This measure is domain-agnostic and requires no interaction data, and its effectiveness here suggests it may serve as a general content-based similarity signal wherever items can be represented as distributions over features. Future research can operationalize this non-acoustic residual directly. First, incorporating reviewer metadata would allow researchers to test whether critics with higher field-specific capital~\cite{corciolani2020} generate edges with larger non-acoustic residuals compared to trade reviewers. Second, combining audio models with other existing metadata such as geography, record label, mutual collaborations, and artist interviews could help explain the sociological remainder (i.e. the portion of artist similarity not explained by audio content). A third and more applied direction follows directly from the cold-start setting: having established that acoustic distance recovers critic adjacency for unseen artists, the predicted adjacency scores can be used to \emph{rank} candidate neighbors for a cold-start artist, and evaluated as a retrieval task against audio-only and behavioral baselines with ranking metrics such as NDCG and recall. This reframes the present construct-validity result as the retrieval-side foundation of a unified search-and-recommendation setting, in which an interpretable, privacy-preserving critic signal is combined with content and interaction signals rather than used in isolation; validating that the signal is acoustically grounded is a prerequisite for such joint models. Ultimately, using audio features as an independent physical baseline allows recommender systems to leverage critical discourse responsibly, using higher-consensus edges for robust cold-start discovery while recognizing single-source commentary as a rich record of cultural and narrative history. In doing so, this work completes a three-part case for critical adjacency as a first-class signal for recommendation and discovery: it is theoretically motivated by the sociology of criticism, internally valid as a graph, coherent under community detection and effective in recommendation simulation~\cite{badillo-goicoechea-hdsr}, and, as shown here, externally grounded in the acoustic content of the music itself.

\section*{Ethics and Privacy Statement}
This work uses publicly available audio descriptors and a graph derived from publicly available music publications; it involves no personal data or human subjects. A content-based discovery signal of this kind could reinforce existing coverage biases if deployed uncritically, since both its audio corpus and its critical target over-represent Anglophone and canonical artists.


\begin{thebibliography}{99}

\bibitem{adomavicius2010}
Gediminas Adomavicius and Alexander Tuzhilin. 2010. Context-Aware Recommender Systems. \emph{Recommender Systems Handbook} (2010), 217--253.

\bibitem{badillo-goicoechea-hdsr}
Elena Badillo-Goicoechea. 2025. Modeling Artist Influence for Music Selection and Recommendation: A Purely Network-Based Approach. \emph{Harvard Data Science Review} 7, 4. https://doi.org/10.1162/99608f92.fb935f61

\bibitem{berenzweig2004}
Adam Berenzweig, Beth Logan, Daniel P. W. Ellis, and Brian Whitman. 2004. A Large-Scale Evaluation of Acoustic and Subjective Music-Similarity Measures. \emph{Computer Music Journal} 28, 2 (2004), 63--76. https://doi.org/10.1162/014892604323112257

\bibitem{bogdanov2016}
Dmitry Bogdanov, Alastair Porter, Perfecto Herrera, and Xavier Serra. 2016. Cross-collection Evaluation for Music Classification Tasks. In \emph{Proc. ISMIR}.

\bibitem{bourdieu1984}
Pierre Bourdieu. 1984. \emph{Distinction: A Social Critique of the Judgement of Taste}. Harvard University Press.

\bibitem{bourdieu1993}
Pierre Bourdieu. 1993. \emph{The Field of Cultural Production: Essays on Art and Literature}. Columbia University Press.

\bibitem{cohen2000web}
William W. Cohen and Wei Fan. 2000. Web-collaborative Filtering: Recommending Music by Crawling the Web. \emph{Computer Networks} 33, 1--6 (2000), 685--698.

\bibitem{corciolani2020}
Matteo Corciolani, Kent Grayson, and Ashlee Humphreys. 2020. Do More Experienced Critics Review Differently? How Field-Specific Cultural Capital Influences the Judgments of Cultural Intermediaries. \emph{European Journal of Marketing} 54, 3 (2020), 478--510.

\bibitem{deldjoo2024}
Yashar Deldjoo, Markus Schedl, and Peter Knees. 2024. Content-driven Music Recommendation: Evolution, State of the Art, and Challenges. \emph{Computer Science Review} 51 (2024), 100618. https://doi.org/10.1016/j.cosrev.2024.100618

\bibitem{ellis2002}
Daniel P. W. Ellis, Brian Whitman, Adam Berenzweig, and Steve Lawrence. 2002. The Quest for Ground Truth in Musical Artist Similarity. In \emph{Proc. ISMIR}, 170--177.

\bibitem{fabbri1982}
Franco Fabbri. 1982. A Theory of Musical Genres: Two Applications. In \emph{Popular Music Perspectives}, 52--81.

\bibitem{flexer2016}
Arthur Flexer and Thomas Grill. 2016. The Problem of Limited Inter-rater Agreement in Modelling Music Similarity. \emph{Journal of New Music Research} 45, 3 (2016), 239--251.

\bibitem{frith1996}
Simon Frith. 1996. \emph{Performing Rites: On the Value of Popular Music}. Harvard University Press.

\bibitem{hirsch1972}
Paul M. Hirsch. 1972. Processing Fads and Fashions: An Organization-Set Analysis of Cultural Industry Systems. \emph{American Journal of Sociology} 77, 4 (1972), 639--659.

\bibitem{javed2021review}
Umair Javed, Kamran Shaukat, Ibrahim A. Hameed, Farhat Iqbal, Talha Mahboob Alam, and Suhuai Luo. 2021. A Review of Content-Based and Context-Based Recommendation Systems. \emph{International Journal of Emerging Technologies in Learning} 16, 3 (2021), 274. https://doi.org/10.3991/ijet.v16i03.18851

\bibitem{knees2013survey}
Peter Knees and Markus Schedl. 2013. A Survey of Music Similarity and Recommendation from Music Context Data. \emph{ACM Trans. Multimedia Comput. Commun. Appl.} 10, 1 (2013), 1--21. https://doi.org/10.1145/2542205.2542206

\bibitem{lena2008}
Jennifer C. Lena and Richard A. Peterson. 2008. Classification as Culture: Types and Trajectories of Music Genres. \emph{American Sociological Review} 73, 5 (2008), 697--718.

\bibitem{lena2012}
Jennifer C. Lena. 2012. \emph{Banding Together: How Communities Create Genres in Popular Music}. Princeton University Press.

\bibitem{lindberg2005}
Ulf Lindberg. 2005. \emph{Rock Criticism from the Beginning: Amusers, Bruisers, and Cool-Headed Cruisers}. Peter Lang.

\bibitem{mcfee2012}
Brian McFee and Gert R. G. Lanckriet. 2012. Hypergraph Models of Playlist Dialects. In \emph{Proc. ISMIR}, 343--348.

\bibitem{mckay2006}
Cory McKay and Ichiro Fujinaga. 2006. Musical Genre Classification: Is It Worth Pursuing and How Can It Be Improved?. In \emph{Proc. ISMIR}, 101--106.

\bibitem{negus1999}
Keith Negus. 1999. \emph{Music Genres and Corporate Cultures}. Routledge.

\bibitem{oramas2015semantic}
Sergio Oramas, Mohamed Sordo, Luis Espinosa-Anke, and Xavier Serra. 2015. A Semantic-Based Approach for Artist Similarity. In \emph{Proc. 16th International Society for Music Information Retrieval Conference (ISMIR)}, 100--106.

\bibitem{peyre2019computational}
Gabriel Peyr{\'e} and Marco Cuturi. 2019. Computational Optimal Transport. \emph{Foundations and Trends in Machine Learning} 11, 5--6 (2019), 355--607. https://doi.org/10.1561/2200000073

\bibitem{radovanovic2009}
Milo{\v s} Radovanovi{\'c}, Alexandros Nanopoulos, and Mirjana Ivanovi{\'c}. 2009. Nearest Neighbors in High-Dimensional Data: The Emergence and Influence of Hubs. In \emph{Proc. 26th ICML}, 865--872.

\bibitem{regev2013}
Motti Regev. 2013. \emph{Pop-Rock Music: Aesthetic Cosmopolitanism in Late Modernity}. John Wiley \& Sons.

\bibitem{schedl2014mir}
Markus Schedl, Emilia G{\'o}mez, and Juli{\'a}n Urbano. 2014. Music Information Retrieval: Recent Developments and Applications. \emph{Foundations and Trends in Information Retrieval} 8, 2--3 (2014), 127--261.

\bibitem{schedl2018}
Markus Schedl, Hamed Zamani, Ching-Wei Chen, Yashar Deldjoo, and Mehdi Elahi. 2018. Current Challenges and Visions in Music Recommender Systems Research. \emph{International Journal of Multimedia Information Retrieval} 7, 2 (2018), 95--116.

\bibitem{schein2002coldstart}
Andrew I. Schein, Alexandrin Popescul, Lyle H. Ungar, and David M. Pennock. 2002. Methods and Metrics for Cold-Start Recommendations. In \emph{Proc. 25th ACM SIGIR}, 253--260. https://doi.org/10.1145/564376.564421

\bibitem{seyerlehner2010}
Klaus Seyerlehner, Gerhard Widmer, and Peter Knees. 2010. A Comparison of Human, Automatic and Collaborative Music Genre Classification. In \emph{International Workshop on Adaptive Multimedia Retrieval}, 118--131.

\bibitem{silver2016}
Daniel Silver, Monica Lee, and C. Clayton Childress. 2016. Genre Complexes in Popular Music. \emph{PLOS ONE} 11, 5 (2016), e0155471. https://doi.org/10.1371/journal.pone.0155471

\bibitem{sturm2023validity}
Bob L. Sturm and Arthur Flexer. 2023. A Review of Validity and Its Relationship to Music Information Research. In \emph{Proc. ISMIR}, 47--55.

\bibitem{vahutre2025}
Nikita Vahutre, Sujata S. Adagale, and Priya A. Kadam. 2025. A Comprehensive Survey of Content-Based Music Recommendation Techniques. In \emph{Business Data Analytics}, 360--368. https://doi.org/10.1007/978-3-031-80778-7\_26

\bibitem{vandenoord2013deep}
A{\"a}ron van den Oord, Sander Dieleman, and Benjamin Schrauwen. 2013. Deep Content-based Music Recommendation. In \emph{Advances in Neural Information Processing Systems}.

\bibitem{vanderlaan2007superlearner}
Mark J. van der Laan, Eric C. Polley, and Alan E. Hubbard. 2007. \emph{Super Learner}.

\bibitem{villani2009optimal}
C{\'e}dric Villani. 2009. \emph{Optimal Transport: Old and New}. Springer Berlin Heidelberg. https://doi.org/10.1007/978-3-540-71050-9

\bibitem{wiggins2009}
Geraint A. Wiggins. 2009. Semantic Gap?? Schemantic Schmap!! Methodological Considerations in the Scientific Study of Music. In \emph{11th IEEE International Symposium on Multimedia}, 477--482.

\end{thebibliography}
\end{document}